\documentclass{article}
\usepackage[utf8]{inputenc}
\usepackage{graphicx}
\usepackage{algorithm}
\usepackage{algpseudocode}
\usepackage[numbers]{natbib}
\usepackage[margin=1in]{geometry}
\usepackage{xcolor}

\newcommand*\samethanks[1][\value{footnote}]{\footnotemark[#1]}
\usepackage{float}

\title{DEFT: Diverse Ensembles for Fast Transfer in Reinforcement Learning}
\author{Simeon Adebola\thanks{Equal Contribution}, Satvik Sharma\samethanks, Kaushik Shivakumar\samethanks}

\DeclareMathDelimiter{\Vert}
  {\mathord}{symbols}{"6B}{largesymbols}{"0D}
\let\|=\Vert

\DeclareMathSymbol{\parallel}{\mathrel}{symbols}{"6B}

\date{}

\begin{document}
\maketitle

\section{Extended Abstract}
Deep ensembles have been shown to extend the positive effect seen in typical ensemble learning to neural networks and to reinforcement learning (RL). However, there is still much to be done to improve the efficiency of such ensemble models. In this work, we present Diverse Ensembles for Fast Transfer in RL (DEFT), a new ensemble-based method for reinforcement learning in highly multimodal environments and improved transfer to unseen environments. The algorithm is broken down into two main phases: training of ensemble members, and synthesis (or fine-tuning) of the ensemble members into a policy that works in a new environment.

The first phase of the algorithm involves training regular policy gradient or actor-critic agents in parallel but adding a term to the loss that encourages these policies to differ from each other. This causes the individual unimodal agents to explore the space of optimal policies and capture more of the multimodality of the environment than a single actor could. The alternative is allowing a single agent to represent multimodal policies, which is often complex, of high variance, and unwieldy to work with.

The second phase of DEFT involves synthesizing the component policies into a new policy that works well in a modified environment, in one of two ways. The first is using a hierarchical policy to select between the component policies, optionally fine-tuning the ensemble members as well. The second is partially supervising the training of a new policy with the closest ensemble member in action-space. We show that both schemes work to fine tune policies, but the latter shows the most promise in speeding up this process significantly. Moreover, the very fact that we train an ensemble of policies (and that each component policy fine-tunes differently) means that practically, one could fine-tune them in parallel and choose the fastest to converge.

To evaluate the performance of DEFT, we start with a base version of the Proximal Policy Optimization (PPO) algorithm and extend it with the modifications for DEFT. We primarily focus on two Mujoco environments: Walker and Ant, which we modify to be more multimodal. Our results show that the pretraining phase is effective in producing diverse policies in multimodal environments. For example, in an environment where Ants are rewarded for running in any direction, the resulting policies run in circles, run in different directions, or perform a combination of the two. Furthermore, especially in the modified Ant environments, DEFT often converges to a high reward significantly faster than alternatives, such as random initialization without DEFT and fine-tuning of ensemble members. 

While there is certainly more work to be done to analyze DEFT theoretically and extend it to be even more robust, we believe it provides a strong framework for capturing multimodality in environments while still using RL methods with simple policy representations.

\newpage
\clearpage
\section{Introduction}

\subsection{Motivation}
Ensemble learning involves combining different models together to make a prediction on input data with the goal of having better, more robust results than any of the individual models could achieve. Popular ensemble models include random forests, bagging, boosting, and stacking \cite{nguyen_ensemble_2020, ganaie2021ensemble}.
Ensemble learning provides a way to reduce the variance that is a known issue with neural networks \cite{brownlee_ensemble_2018} and has been applied to them with positive results \cite{1999}. Ensemble learning can similarly be extended to reinforcement learning (RL), which previous work has shown to be effective \cite{seerl, sunrise, chen2017ucb}.

In this paper, we present our work on deep ensembles for fast transfer in RL with a focus on multimodal environments. 
While there are already existing methods to represent multimodal agent behavior, such as Q learning (specifically soft Q learning \cite{sql}), other methods like soft actor-critic \cite{sac} have become more popular, but do not have a good way of representing multimodality. Our goal is to enable actor-critic style methods to represent multimodality with ensembles, and subsequently leverage ensembles to perform rapid transfer to new environments. By building on current state-of-the-art methods and algorithms, we show high performance on transfer tasks.




Our primary contributions in this work are the following: 
\begin{enumerate}
\item We present a new approach for ensemble learning in RL that allows the use of simple actor representations yet can represent multimodal policies in an environment.
\item We show that our approach works on transfer tasks in multiple Mujoco environments with an emphasis on multimodality. 
\item We discuss our current results and intuition for them: why our approach beats regular fine-tuning in certain environments and not others.
\end {enumerate}
In the next sections, we cover these contributions. The remainder of this section presents related work, section 3 presents our methodology, section 4 showcases our results, and section 5 presents our conclusions and future work.

\subsection{Related Work}

As mentioned before, there has been prior work done related to different aspects of DEFT. Here, we discuss work that is both directly and indirectly related to methods that we present in this paper.
\subsubsection{Ensemble Methods for RL}

Ensemble methods have been used in RL to some success. For example, Sample-efficient ensemble reinforcement learning (SEERL) \cite{seerl} trains policies sequentially, beginning the training of a new policy by significantly increasing the learning rate on a converged policy. After performing this step, it uses a selection step to choose the best policies that are also sufficiently different from one another. At evaluation time, SEERL uses one of many potential methods to combine the top policies to ultimately produce a policy $\pi$. Compared to this work, we work to actively encourage diversity between component policies in our ensemble by using a penalty term. We also design our ensembles to transfer well to new environments.

Another recent technique that makes use of ensembles is SUNRISE \cite{sunrise}, which uses a Q-ensemble to both drive more efficient exploration and reduce the amount of error in Bellman backups through weighting based on the variance in the ensemble estimates.

\subsubsection{Existing Exploration and Soft Methods for Transfer}
There is also much existing work on maximum entropy, or soft, reinforcement learning methods, such as Soft Actor-Critic \cite{sac} and Soft Q-Learning \cite{sql}, both designed to encourage exploration by adding terms that favor the entropy of resulting policies.

Q learning in general becomes more difficult as action spaces become larger and continuous. However, actor-critic methods like SAC are usually implemented with the choice of a Gaussian for the policy action representation. In our work, we attempt to find the best of both worlds by training multiple actor-critic agents, each of which may have unimodal policy representations, and then combining them at test time on a new environment.

\subsubsection{Unsupervised Reinforcement Learning}
We believe that unsupervised RL contains many ideas that can be built upon to improve the ability of RL agents to express multimodality and also transfer well to new environments.

In general, methods designed to learn useful behaviors without reward use proxies like entropy as signals for exploration. For example, the Diversity is All You Need \cite{Diversity} paper learns useful behavior (as various skills) without a reward function. The work explores how these skills can be used for fine-tuning or hierarchical RL when a reward function becomes available. We explore many similar ideas to this paper but in a fully supervised setting, and we also explore different methods to obtain diverse policies and fine-tune on environments using these.

\subsubsection{Policy Distillation for Transfer}
There has also been work on distilling policies for transferring to new domains \cite{transfer}. This type of procedure usually involves one or more teacher agents, whether they are policies or Q functions, being used to supervise a student agent and is often used in the context of multi-task learning.

Our work differs from this significantly since we control the training of the teacher policies to be purposefully diverse, we seek to minimize the distance from the \textit{closest}, not \textit{all} teacher policies, and we minimize this distance while fine-tuning the policy on a target domain. 

\subsubsection{Our Work}
Our approach combines ideas from a variety of these works to propose a new ensemble-based method for reinforcement learning and transfer to unseen environments with minimal fine-tuning. One way to view our work is as an algorithm that is in between Q-Learning and Actor-Critic methods; allowing multimodality to be represented while still retaining favorable inductive bias due to the simplicity of policy representations. Another way of viewing DEFT is as an adaptation and modification of unsupervised RL methods to exploit multimodality in policies in supervised settings.

\section{Methods}
DEFT does not use ensembles in the traditional sense, or for the traditional purpose, where all the neural networks are trained to solve the same task independently from each other. We also differentiate ourselves from the traditional methods of reducing ensemble variance by inducing slight biases through bagging or the use of feature subsets in random forests \cite{randomforestsubset}. However, we do indeed induce bias into the training procedure in the hopes of improving diversity. DEFT can generally be divided into two stages, training of the component policies and adaptation to new environments.

\subsection{Training of the Component Policies}
The goal of the training phase is to produces policies $\pi_e$ that all attempt to maximize the reward in an environment, while also being significantly different from each other. If the policies produce Gaussian distributions for actions conditioned on observations (in continuous action spaces), given fixed standard deviations, the KL divergence between them is approximately affine in the mean squared error (MSE) between the two policies. Taking inspiration from this, we define a policy divergence terms as follows:

\begin{equation}
    d_{\pi_i, \pi_j} = E_{\tau \sim \pi_i(\tau)} [c(|| E[\pi_j(a|o)] - E[\pi_i(a|o)] ||^2)]
\end{equation}

\begin{equation}
    d_{\pi_i, \pi_j}(o) = [c(|| E[\pi_j(a|o)] - E[\pi_i(a|o)] ||^2)]
\end{equation}

where $c$ is a concave function that prevents the gradients resulting from vastly different policies from overpowering the REINFORCE objective at any time. The second term is conditioned on observations as well.

In many of our experiments, we choose to use the sigmoid function $\sigma$ for $c$, but there are likely many other choices that would have sufficed. The intuition for this term is that as long as policies are different beyond a certain threshold, it does not matter as to \textit{how different} they are.

Note that the expectation is taken under the distribution of trajectories resulting from the $\pi_i$, so this term is not commutative. We train our policies in parallel with each other, such that they simultaneously try to maximize their difference from other policies and their reward. Based on this intuition, we define a modified global objective function (to be maximized) for the training of the policies and a modified loss for the policy updates:

\begin{equation}
    J(\theta) = \sum_{i \in N_E} (E_{\tau \sim \pi_{\theta_i}(\tau)} [r(\tau)] + \delta E_{j \sim \mathcal{U}([0, N_E-1] \setminus i)}[d_{\pi_{\theta_i}, \pi_{\theta_j}}]) 
\end{equation}

where $N_E$ is the number of ensemble members. The way we maximize this objective is by modifying the loss function for the individual ensemble members. We define

\begin{equation}
    L^{div}_{\pi_i} = E_{j \sim \mathcal{U}([0, N_E-1] \setminus i)}[d_{\pi_{\theta_i}, \pi_{\theta_j}}]
\end{equation}

The total loss used to update the reinforcement algorithm''s policy is the following, where $\delta$ is a hyperparameter used to control the amount of differentiation between the policies.

\begin{equation}
    L_{\pi_i} = L^{orig}_{\pi_i} - \delta L^{div}_{\pi_i}
\end{equation}
We backpropagate only into the current policy $\pi_i$ when updating weights. Finally, it is worth noting that alternate schemes such as bonus rewards for divergence of the state distributions rather than purely policy distributions may also work well for certain domains. Because the first and second phases are isolated, there are likely other permissible schemes for encouraging diversity.

\subsection{Fine-tuning on New Environments}
\subsubsection{Using a Hierarchical Policy}
The hierarchical policy, $\Lambda$, is trained after the ensemble members have been trained on the original environment. The hierarchical policy is trained with a set of ensemble members, $$ S = \{\pi_{i} | i \in [0, n-1]\} $$ Thus, the action space is discrete with a size of $N_E$. During training, the hierarchical policy selects an ensemble member, $\pi_{i}$, from set $S$ and uses $\pi_{i}$ to select action $a = \pi_{i}(a_{t}|o_{t})$ given the observation $o_{t}$ seen at time $t$ in the environment. The rest of the training procedure is trained in a manner similar to standard PPO.

\subsubsection{Using Partial Supervision for Fine-tuning}
One potential problem with hierarchical policies is that they may be vastly underparametrized or overparametrized, based on whether the component policies are allowed to be fine tuned or not. An alternative technique that we develop for transferring to new environments using the ensemble involves not necessarily reusing parts of the ensemble members like in the hierarchical policy, but instead training an entirely separate policy with supervision from the ensemble members. The key insight is that we can vastly reduce the number of parameters needed for a policy representation but still be able to use the ensemble members in a meaningful way to improve sample efficiency. Furthermore, we could use an arbitrarily different architecture and even modified state spaces (see Results section) as long as the ensemble policies also produce actions given observations.

The main question to address then becomes how we choose \textit{which} of the ensemble members to supervise the new policy $\pi'$ with. The choice we make is the ensemble policy $\pi_e$ with the \textit{lowest prediction error} for $\pi'$. Intuitively, this means that if $\pi'$ is tending towards behaviors that ensemble members are already proficient at, the ensemble members can accelerate the learning of these behaviors through a certain amount of supervision. 

Specifically, we have $$\Pi = \{\pi_0, \pi_1, ...\}, |\Pi| = N_E$$ as usual after our diversity-based pretraining step. These are the policies trained on an initial environment, designed to capture useful behaviors for the target environment. We then train a new policy $\pi'$ but add a term to the loss in order to partially supervise it so that it adheres more closely to the component policies. This can be seen as biasing the training in favor of policies that are close to at least one of the ensemble methods, in turn lowering the variance. In contrast to other methods of transfer learning for reinforcement learning, our method does not necessarily require a non-random initializing of its weights as this supervision process happens online, over the course of training.

The RL objective for this second step of training is modified to the following, where $\rho$ controls the magnitude of the partial supervision term.
\begin{equation}
    J(\theta) = E_{\tau \sim \pi_{\theta'}(\tau)} [r(\tau)] - \rho \min_{j \in [0, N_E - 1]} d_{\pi_{\theta_i}, \pi_{\theta_j}}
\end{equation}

\begin{figure}
    \centering
    \includegraphics[width=1.0\linewidth]{"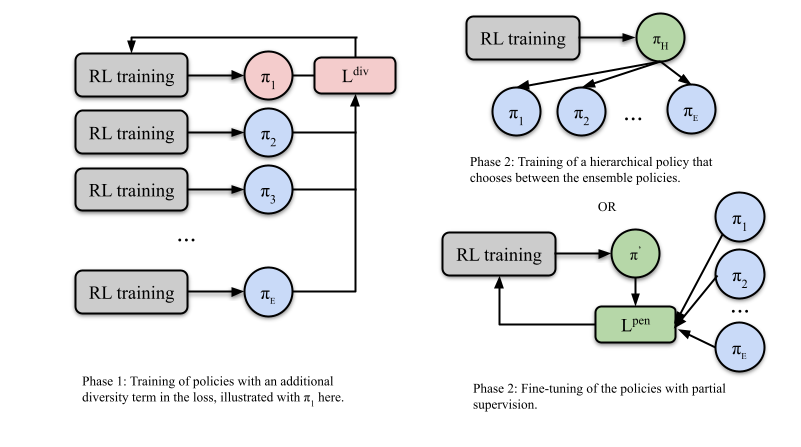"}
    \caption{An overview of the DEFT pipeline.}
    \label{fig:DEFT}
\end{figure}

Thus, the total loss we use for training the new policy is as follows.

$$L = L_{orig} + \rho \min_{j \in [0, N_E - 1]} d_{\pi_{\theta_i}, \pi_{\theta_j}}$$

where c (within the $d$ term) is again a concave function to avoid penalizing the policy too much if it is very far from the ensemble members, indicating that the new environment needs a $\pi'$ significantly different from any of the $\pi_e$. The DEFT algorithm is summarized in Algorithm \ref{alg:cap} and Figure \ref{fig:DEFT}.

\subsubsection{Parallel Fine-Tuning and Optimal Policy Selection}
The very fact that an ensemble of policies exists (and that each component policy fine-tunes differently) means one could fine-tune them in parallel and choose the fastest to converge. In other words, a random fine-tuned policy from an ensemble will underperform the best fine-tunable member of the ensemble. This means that the presence of an ensemble as opposed to a single policy is itself a contribution that results in faster transfer.

\begin{algorithm}
\caption{DEFT Algorithm (using partial supervision for fine-tuning). Note that the usual RL algorithm's instructions and updates are omitted for brevity to highlight the differences.}\label{alg:cap}
\begin{algorithmic}
\Require Original environment $E_{orig}$ \Comment{Pre-training}
\State $\Pi \gets \{\pi_{\theta_1}, \pi_{\theta_2}, ..., \pi_{\theta_{N_E}}\}$ with random initialization
\While{$t < T$}
    \For{$\pi_{\theta_i} \in \Pi$}
        \State Original steps before policy update
        \State $L_{orig} \gets$ regular policy loss
        \State $j \sim \mathcal{U}([0, N_E-1] \setminus i)$
        \State $L_{div} \gets d_{\pi_{\theta_i}, \pi_{\theta_j}}$
        \State $L \gets L^{orig} - \delta L_{div}$
        \State $\theta_i \gets \theta_i - \alpha \nabla_{\theta_i} L$
        \State Original steps after policy update
    \EndFor
\EndWhile

\Require New environment $E_{new}$, $\Pi = \{\pi_0, \pi_1, ...\}, |\Pi| = N_E$ \Comment{Fine-tuning}
\State $\pi_{\theta'} \gets$ \{random, oneof, avg\} initialization
\While{$t < T'$}
    \State Original steps before policy update
    \State $L_{orig} \gets$ regular RL algorithm policy loss
    \State $L_{pen}(o) \gets \min_{i \in [0, N_E - 1]} d_{\pi_{\theta'}, \pi_{\theta_i}}(o)$ \Comment{$o$ is a batch of observations}
    \State $L \gets L_{orig} + \frac{\rho}{\mathcal{|O|}} \sum_{o} {L_{pen} (o)}$
    \State $\theta' \gets \theta' - \alpha \nabla_{\theta'} L$
    \State Original steps after policy update
\EndWhile

\State{$\pi_{eval} \gets \pi_{\theta'}$}
\end{algorithmic}
\end{algorithm}

\section{Results}
We base our experiments off of Proximal Policy Optimization (PPO) \cite{ppo}. We choose this algorithm as it runs quickly and works well on continuous control tasks. Our work can be easily applied to any other RL algorithm that makes use of a neural network to estimate $\pi$, whether it is a regular policy gradient method or an actor-critic method.

There are five main environments of interest that we use to study transfer. We use three Ant environments based on Mujoco's Ant-v3. The first, $A_o$, used for pretraining, is modified to increase its multimodality, giving equal reward to agents for running in any direction. For transfer, we try two environments: $A_{dir}$, which rewards an agent's velocity specifically in the $(\sqrt{2}/2, \sqrt{2}/2)$ unit vector direction, and $A_{maze}$, where the goal is to run as fast to the left as possible, given a simple maze (with negative rewards acting as walls) requiring the Ant to run up briefly before running to the left. We base the Walker experiments on a Walker-v3 environment, where the modified version has significantly less friction and thus has a tendency to slip and fall. The original Walker environment and the lower friction modified environment are referred to as $W_o$ and $W_m$ respectively. We also conduct some tests on the Pointbot environment which are included in the appendix. 

\subsection{Phase 1, Diversity-Promoting Pre-training}
\subsubsection{Ant Environment}
\begin{figure}[H]
    \centering
    \includegraphics[width=0.6\textwidth]{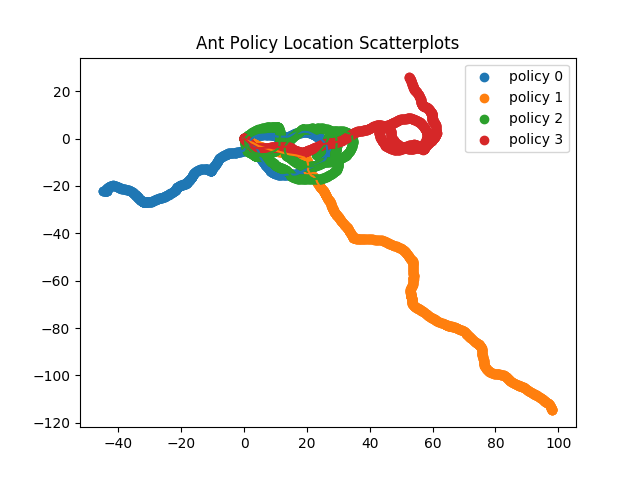}
    \caption{Ant agents which are trained with diversity terms. As shown by this figure, it is apparent that they are exploiting the multimodality of optimal policies for the environment. Each of the agents travels in a different combination of circles and lines in different directions.}
    \label{fig:antsrunning}
\end{figure}

Four Ant agents are trained in $A_{o}$ with $\delta=0.002$. Figure \ref{fig:antsrunning} illustrates that these agents learn different behaviors: not merely running straight in different directions, but one agent actually runs in circles. This figure provides great insights into the way in which the diversity term in the pretraining phase helps to exploit the multimodality of the environment. Purely state-based exploration may not adequately reward the different behaviors.

\subsubsection{Walker Environment}
\begin{figure}[H]
    \centering
    \includegraphics[width=0.35\textwidth]{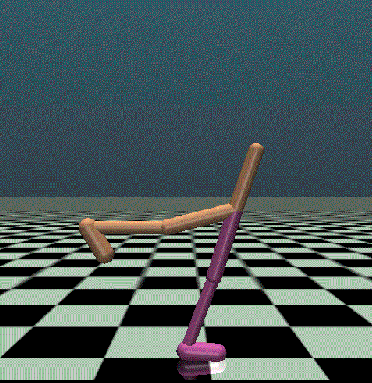}
    \includegraphics[width=0.37\textwidth]{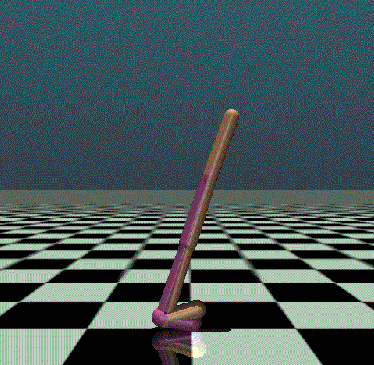}
    \caption{\textbf{Left}: A frame of the first agent learning to use one leg to hop and run to maximize the reward \textbf{Right}: A frame of the second agent learning to use both legs to hop and run.}
    \label{fig:walkersrunning}
\end{figure}

Two Walker agents are trained in $W_{o}$. These agents learn slightly different behaviors but both accomplish the task. Figure \ref{fig:walkersrunning} shows freeze frames from both agents, illustrating how one learns to separate its legs and the other learns to perform more of a hopping action.

\subsection{Phase 2, Hierarchical}

\subsubsection{Ant Environment}
\begin{figure}[H]
    \centering
    \includegraphics[width=0.45\textwidth]{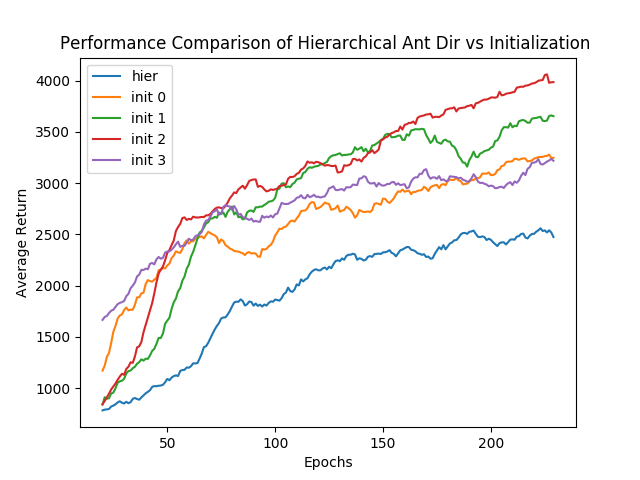}
    \caption{This figure shows that while the hierarchical policy does fine tune reasonably in the target environment, its performance lags behind the fine tuning of individual policies.}
    \label{hierant}
\end{figure}

We train a hierarchical policy over policies trained in $A_o$, in the $A_{dir}$ environment. For this experiment, all $\pi_i \in \Pi$ remain fixed. Figure \ref{hierant} illustrates the performance of this fine-tuning. As can be seen, the inability to fine-tune the component policies likely causes the less effectual fine-tuning than the individual policies.

\subsubsection{Walker Environment}
\begin{figure}[!ht]
    \centering
    \includegraphics[width=0.45\textwidth]{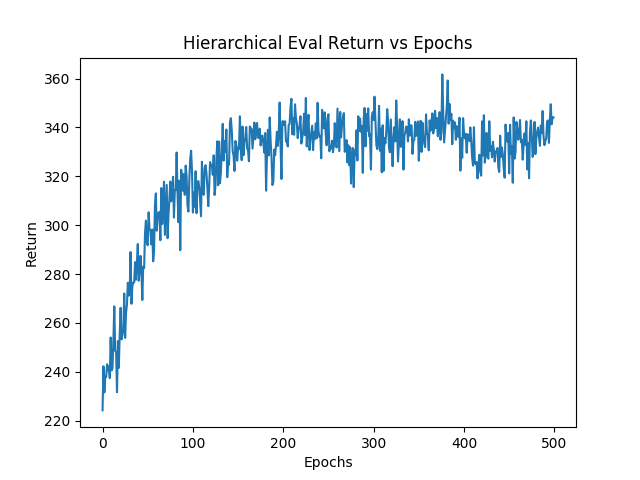}
    \includegraphics[width=0.45\textwidth]{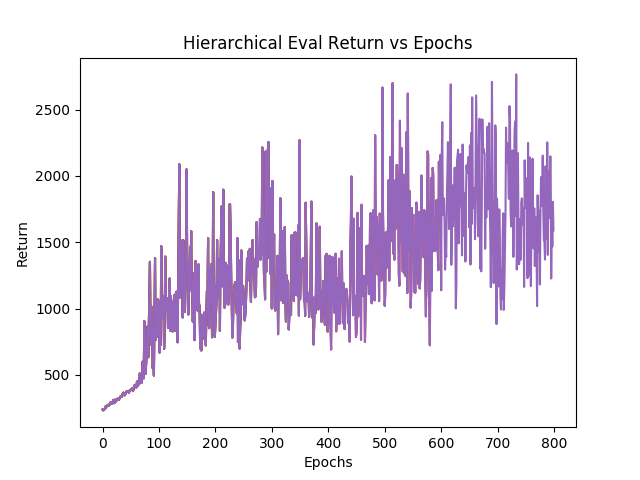}
    \caption{\textbf{Left:} The graph on the left shows the hierarchical policy trained on $W_o$ then fine-tuned on $W_m$ without fine-tuning of the ensemble members. \textbf{Right:} The graph on the right shows first training the ensemble members and the hierarchical policy on $W_o$ then updating the parameters of those policies with $W_m$ for 800 epochs.}
    \label{walker1}
\end{figure}

\begin{figure}[!ht]
    \centering
    \includegraphics[width=0.45\textwidth]{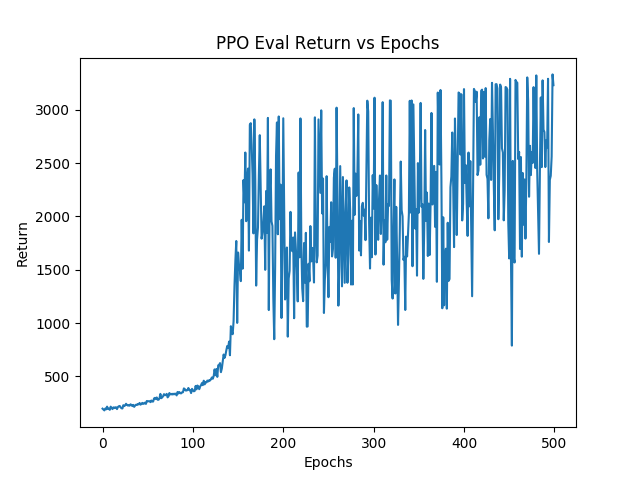}
    \includegraphics[width=0.45\textwidth]{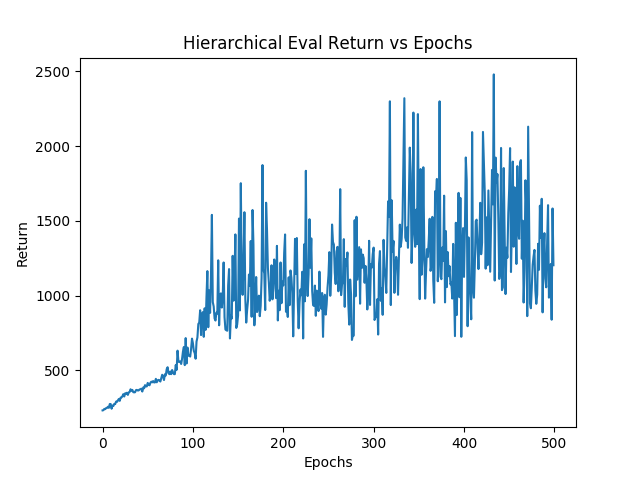}
    \caption{\textbf{Left:} The graph on the left shows standard PPO run first on $W_o$ then fine-tuned on $W_m$. \textbf{Right:} The graph on the right shows first training the ensemble members and the hierarchical policy on $W_o$ then updating the parameters of those policies with $W_m$. Similar hyperparameters were used for both experiments.}
    \label{walker2}
\end{figure}

First, we train two ensemble members and a hierarchical policy on $W_o$ and only fine-tune this hierarchical policy on $W_m$. This does not work well as the agent does not exceed 400 reward as seen in figure \ref{walker1}. Intuitively, each ensemble member is trained with $W_o$; thus choosing any one of them does not allow the hierarchical policy to perform well in $W_m$. This motivates also fine-tuning the corresponding ensemble members simultaneously with the hierarchical policy as seen in both the right purple curve in figure \ref{walker1} and the right blue curve in figure \ref{walker2}. This method is largely unsuccessful for the $W_m$ environment because fine-tuning the hierarchical policy and the corresponding ensemble members creates an overparameterized model which suffers from high variance. One seed of this method is represented by the right graph in figure \ref{walker2} and another in \ref{walker1}. Also in figure \ref{walker2}, the left graph shows a baseline for our method, which is a standard PPO policy fine-tuned by first training that policy in $W_o$ and then updating the parameters of that policy with data from $W_m$. Simultaneously training the hierarchical policy and the ensemble members leads to marginally better performance in the initial phase of training as the hierarchical policy is able to exceed 1000 reward within 110 epochs while the standard PPO policy takes around 150 epochs. 

Since the hierarchical policy has access to multiple ensemble members that are trying to maximize reward on $W_m$, the hierarchical policy can find a particular ensemble member that is performing especially well, and then prefer that member. Thus, the hierarchical policy is able exploit this rapid exploration by the ensemble members to achieve higher reward faster than the standard PPO policy. However, this policy is not able to outperform the standard PPO policy in the long term. We also increase the epochs to 800 to confirm the hierarchical policy does not approach the performance of the PPO policy as seen from the right purple curve in figure \ref{walker1}. However, this purple curve in addition to the right blue curve in figure \ref{walker2} do show that the hierarchical policy consistently learns slightly faster than standard PPO. We see that the hierarchical policy is switching between the ensemble members very quickly, which indicates that the hierarchical policy is not learning how to properly leverage the ensemble members when it is being fine-tuned.

The previous experiment indicates that the hierarchical policy is not able to effectively learn when the rewards can not describe useful behavior for ensemble members to have, such as countering the effect of low friction in $W_m$. In the Ant experiment, when we train a hierarchical policy for a modified environment where the highest reward comes from moving diagonally, it is able to properly learn to move in a particular direction by using its ensemble policies as a basis. Thus, this method can be moderately successful if the rewards are defined broadly enough to reflect the desired performance for ensemble members in possible transfer situations.

\subsection{Phase 2, Partial Supervision}
Most of the experiments for partially supervised transfer are done in the Ant environment.

\subsubsection{Ant Environment}

\begin{figure}[!ht]
    \centering
    \includegraphics[width=0.6\textwidth]{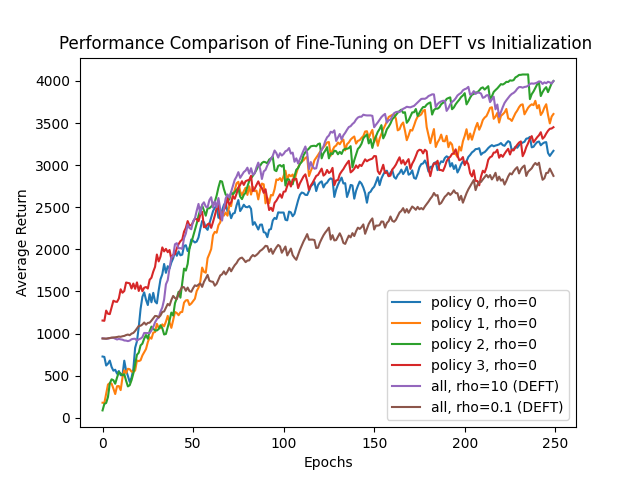}
    \caption{Performance of policies when trained to run in a particular direction for the $A_{dir}$ environment. In this example, $c(x) = x$ rather than the usual choice of sigmoid function. Despite there being significant variance in how the different ensemble members fine-tune, DEFT performs approximately as well as the best of them. Moreover, the true baseline for this experiment is how one of the policies chosen at random performs when fine-tuned, so DEFT certainly outperforms that.}
    \label{fig:ft1}
\end{figure}

As figure \ref{fig:ft1} shows, when the value of $\rho$ is chosen correctly, DEFT does approximately as well as the best of the $\pi_e$ when fine-tuned. In fact, this is exactly the behavior we would expect. In general, transfer learning via fine-tuning will take time to explore and fine-tune the weights sufficiently to achieve good reward on this task, and DEFT naturally does as well as the best (or most relevant) component policy when it is fine-tuned, as it appears to . For these experiments, DEFT is initialized with mean weights as such: $\theta' = \frac{1}{N_E} \sum_{i \in [0, N_E - 1]} \theta_i$.

\begin{figure}[!ht]
    \centering
    \includegraphics[width=0.48\textwidth]{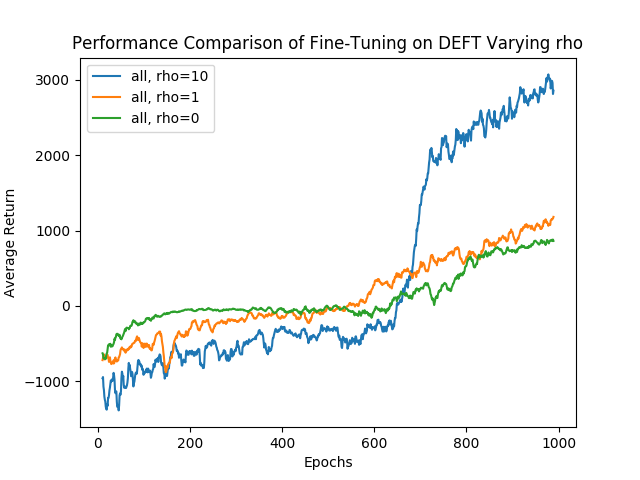}
    \includegraphics[width=0.48\textwidth]{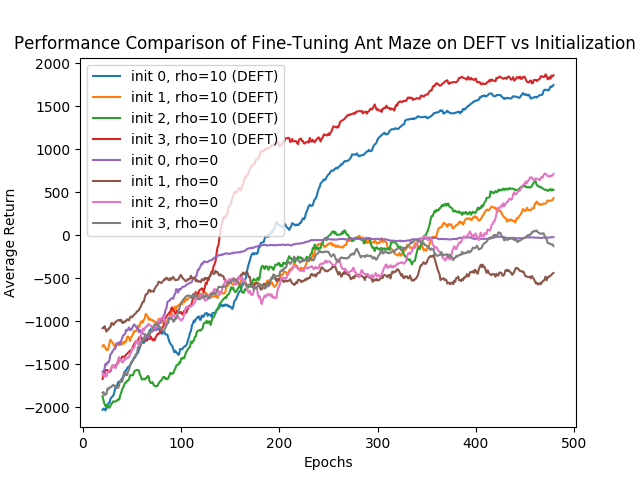}    
    \caption{DEFT on the $A_{maze}$ environment. \textbf{Left:} Comparison of DEFT training with various values of $\rho$, averaged across two runs. This graph clearly shows the benefits of adding the partial supervision from the existing policies. All of the runs here used random initialization. It is not possible to use a pretrained initialization here because the Ant component policies were trained without (x, y) coordinates which are however present in the state vector for this maze environment. \textbf{Right:} Comparison of DEFT (initialized to the starting weights of particular ensemble members) to fine-tuning of individual ensemble members. For two of the initializations, DEFT performs significantly better, whereas for the other two, it performs comparably to fine-tuning.}.
    \label{fig:supervision_helps}
\end{figure}

The second test case uses the $A_{maze}$ environment. As shown in the left part of figure \ref{fig:supervision_helps}, the additional partial supervision helps the policy's return skyrocket at a point; we hypothesize that this is the point where PPO begins exploiting the infinite corridor leftwards, at which point DEFT can begin leveraging the component policies to run leftwards.

For the second part of this experiment, four new component policies including position information were trained on $A_{dir}$ so they could readily be re-initialized in $A_{maze}$. The right half of figure \ref{fig:supervision_helps} illustrates that for two of the initializations, DEFT does far better, and for the others, it performs similarly. Average initialization is not shown, but it did not perform significantly better than the component policies. Overall, it is noteworthy that DEFT is able to stitch together the most relevant policies from a pretraining environment to perform well in a new target domain. The individual ensemble members are not able to perform as well even when fine-tuned because likely their training is not favorably biased in the way DEFT is.

The sensitivity to initializations should be explored further, but it is likely due to the fact that $DEFT$ when tuning does not explore differently from the RL algorithm it is built on. Thus, it may have gotten stuck in a local minimum for some time. We address potential future improvements in the conclusion of this paper.

\subsubsection{Walker Environment}
\begin{figure}[!ht]
    \centering
    \includegraphics[width=0.6\textwidth]{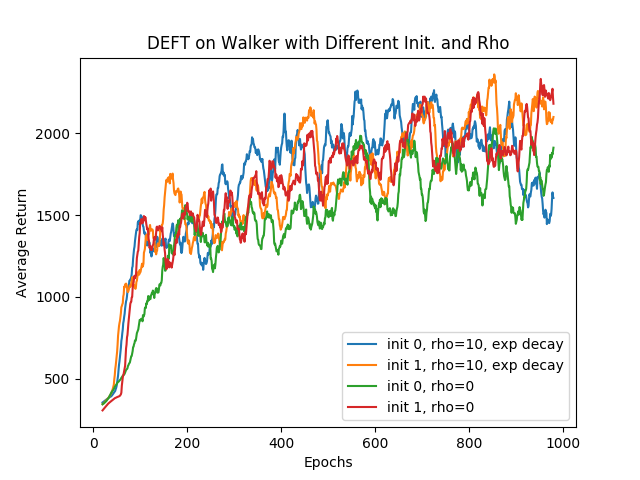}
    \caption{Comparison of DEFT training with various values of $\rho$, and initial policies. In this case, $\rho$ is set to decay from its initial value exponentially over epochs. }
    \label{fig:walker_deft}
\end{figure}

We also investigate the performance of partial supervision on other environments, including the aforementioned $W_m$ after training on $W_o$. As shown in figure \ref{fig:walker_deft}, we find that in the $W_o$ environments, the initial policies do not sufficiently capture the behaviors useful to the $W_m$ environment. Because of this, we find that the performance is not superior to fine-tuning starting from the individual policies, but not worse either.

\section{Conclusions and Future Work}
In this paper, we presented DEFT, which allows the training of diverse ensembles of policies as well as the refining of their performance on new, unseen environments. We do this by adding a diversity term to the training of ensemble members, which we perform in parallel. In the fine-tuning phase, we experiment with both hierarchical policies and partial supervision, the latter of which appears to show more promise. We showcase the performance of DEFT on various environments such as Ant and Walker, highlighting its strengths and weakness as opposed to traditional fine-tuning.

One avenue that may be interesting to explore is performing Cross-Entropy Method (CEM) with the policies in the ensemble as we train them, as well as adjusting the number of policies. Another technique we look forward to experimenting with is not always supervising actions with the closest policy, but also factoring in the uncertainty of the policies to account for distribution shift. The intuition behind this is that we do not wish to supervise our fine-tuned policy with policies that are extremely off-distribution. Thus, we plan to factor in state density estimation to counter this issue to further improve the performance of DEFT. Yet another avenue we may explore is introducing stochasticity so that the closest policy is not always the one selected.

Given that our methods appear to work well despite some sensitivity to hyperparameters, we feel that another interesting line of work would be to understand if we can use meta-learning to fit these parameters. With more time, we would also test these methods with more random seeds as well as on a wider range of environments, which was difficult for us due to runtime considerations. We plan to apply these techniques to the manipulation of deformable objects.

Lastly, our baseline tests show that SAC fine-tunes well (see Appendix). It would be interesting to see how our method could build and improve upon SAC as we have done with PPO. We would explore this in future work. 

\bibliographystyle{plain}
\bibliography{refs}

\begin{thebibliography}{10}

\bibitem{brownlee_ensemble_2018}
Jason Brownlee.
\newblock Ensemble {Learning} {Methods} for {Deep} {Learning} {Neural}
  {Networks}, December 2018.

\bibitem{chen2017ucb}
Richard~Y. Chen, Szymon Sidor, Pieter Abbeel, and John Schulman.
\newblock Ucb exploration via q-ensembles, 2017.

\bibitem{Diversity}
Benjamin Eysenbach, Abhishek Gupta, Julian Ibarz, and Sergey Levine.
\newblock Diversity is all you need: Learning skills without a reward function.
\newblock {\em CoRR}, abs/1802.06070, 2018.

\bibitem{ganaie2021ensemble}
M.~A. Ganaie, Minghui Hu, M.~Tanveer*, and P.~N. Suganthan*.
\newblock Ensemble deep learning: A review, 2021.

\bibitem{sql}
Tuomas Haarnoja, Haoran Tang, Pieter Abbeel, and Sergey Levine.
\newblock Reinforcement learning with deep energy-based policies.
\newblock {\em CoRR}, abs/1702.08165, 2017.

\bibitem{sac}
Tuomas Haarnoja, Aurick Zhou, Pieter Abbeel, and Sergey Levine.
\newblock Soft actor-critic: Off-policy maximum entropy deep reinforcement
  learning with a stochastic actor.
\newblock {\em CoRR}, abs/1801.01290, 2018.

\bibitem{randomforestsubset}
Tin~Kam Ho.
\newblock The random subspace method for constructing decision forests.
\newblock {\em IEEE Transactions on Pattern Analysis and Machine Intelligence},
  20(8):832--844, 1998.

\bibitem{sunrise}
Kimin Lee, Michael Laskin, Aravind Srinivas, and Pieter Abbeel.
\newblock {SUNRISE:} {A} simple unified framework for ensemble learning in deep
  reinforcement learning.
\newblock {\em CoRR}, abs/2007.04938, 2020.

\bibitem{nguyen_ensemble_2020}
Ngoc Minh~Tu Nguyen.
\newblock Ensemble {Reinforcement} {Learning}, April 2020.

\bibitem{1999}
D.~Opitz and R.~Maclin.
\newblock Popular ensemble methods: An empirical study.
\newblock {\em Journal of Artificial Intelligence Research}, 11:169–198, Aug
  1999.

\bibitem{seerl}
Rohan Saphal, Balaraman Ravindran, Dheevatsa Mudigere, Sasikanth Avancha, and
  Bharat Kaul.
\newblock {SEERL:} sample efficient ensemble reinforcement learning.
\newblock {\em CoRR}, abs/2001.05209, 2020.

\bibitem{ppo}
John Schulman, Filip Wolski, Prafulla Dhariwal, Alec Radford, and Oleg Klimov.
\newblock Proximal policy optimization algorithms.
\newblock {\em CoRR}, abs/1707.06347, 2017.

\bibitem{transfer}
Zhuangdi Zhu, Kaixiang Lin, and Jiayu Zhou.
\newblock Transfer learning in deep reinforcement learning: {A} survey.
\newblock {\em CoRR}, abs/2009.07888, 2020.

\end{thebibliography}

\section{Appendix}

\subsection{SAC Baselines}
For our experiments, we primarily build DEFT on top of Proximal Policy Optimization (PPO) \cite{ppo} but we compare with Soft Actor Critic (SAC) \cite{sac} baselines as well, as this algorithm is an instance of a maximum entropy algorithm that is designed to perform better exploration. However, we are aware that too much weight should not be placed on this comparison as PPO is an on-policy algorithm, and SAC is not.

We train SAC on our Ant ($A_o$, $A_{dir}$) and Walker ($W_o$, $W_m$) environments (both original and modified) to get baselines. The results are shown in the figures below. We see that SAC fine-tunes rapidly due to its exploration properties. We mention in our conclusions that we would like to see how well DEFT performs when implemented on top of SAC.

\begin{figure}[!ht]
    \centering
    \includegraphics[width=0.5\linewidth]{"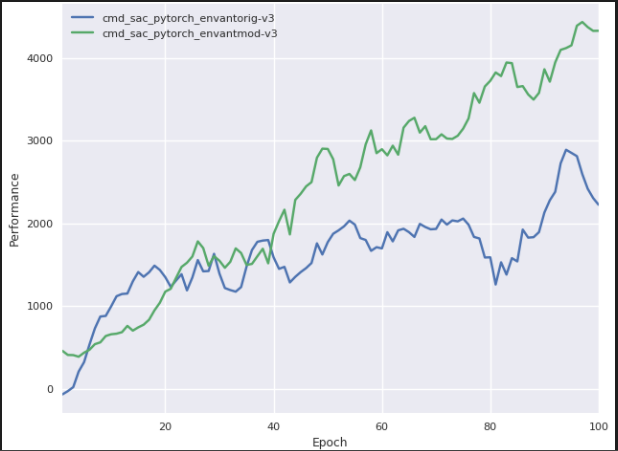"}
    \caption{SAC trained on the modified ant and original ant environments.}
    \label{fig:SAC1}
\end{figure}

\begin{figure}[!ht]
    \centering
    \includegraphics[width=0.5\linewidth]{"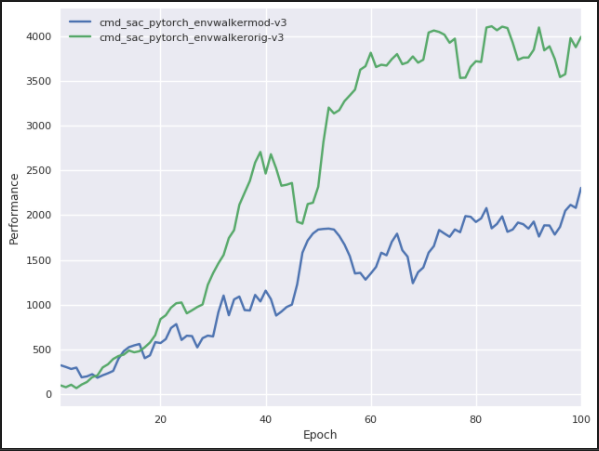"}
    \caption{SAC trained on the modified walker and original walker environments.}
    \label{fig:SAC2}
\end{figure}

\subsection{Pointbot Environment}
\begin{figure}[!ht]
    \centering
    \includegraphics[width=0.48\linewidth]{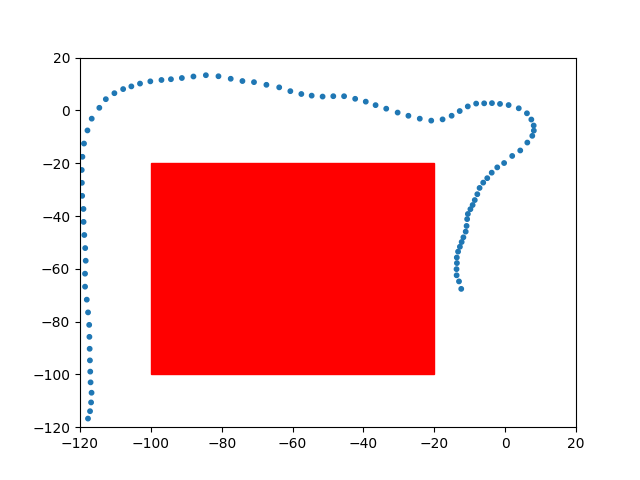}
    \includegraphics[width=0.48\linewidth]{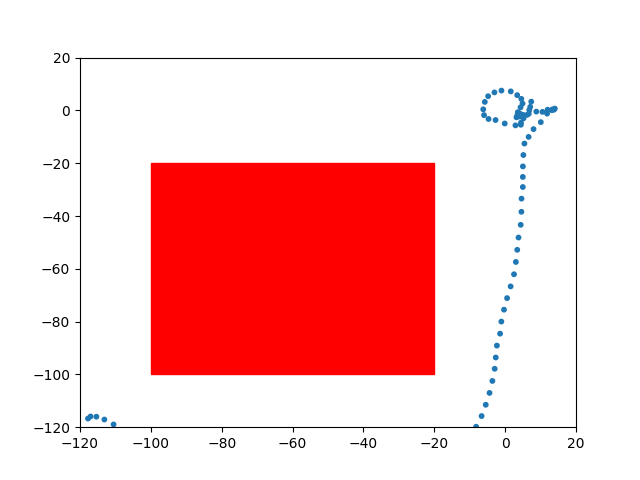}
    \includegraphics[width=0.48\linewidth]{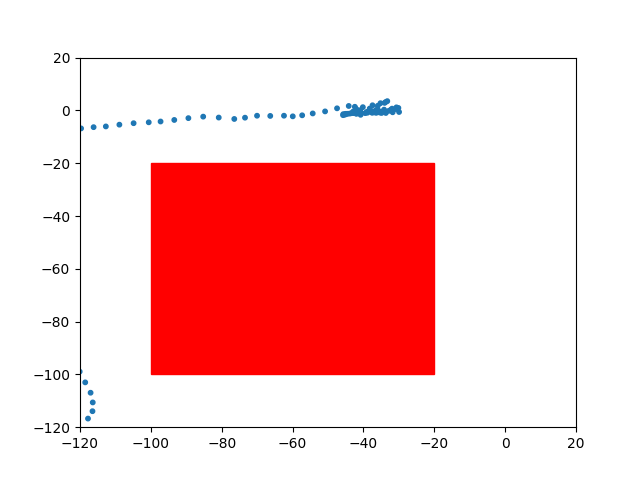}
    \includegraphics[width=0.48\linewidth]{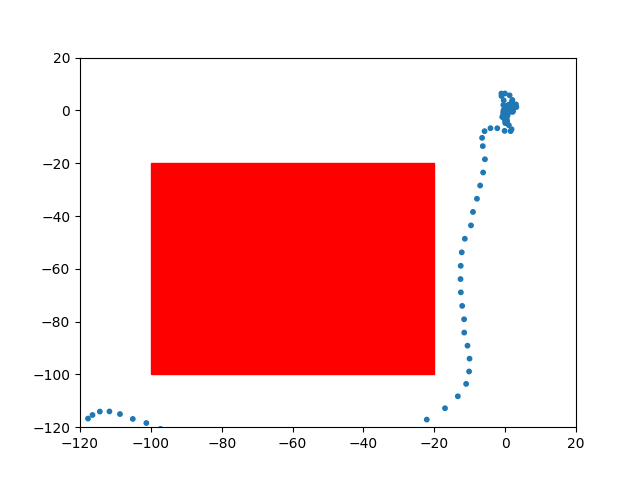}
    \includegraphics[width=0.48\linewidth]{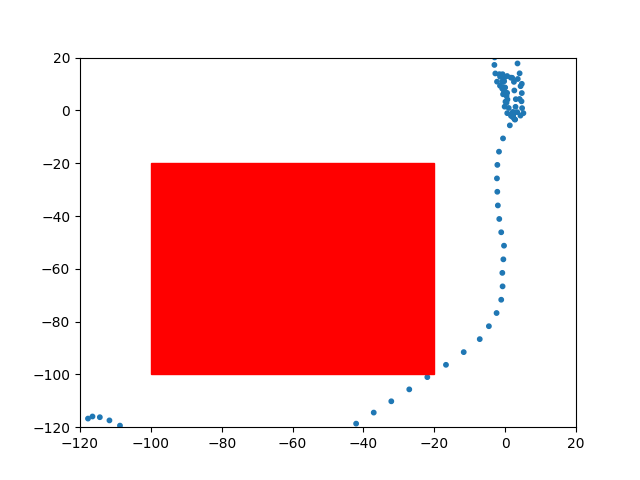}
    \includegraphics[width=0.48\linewidth]{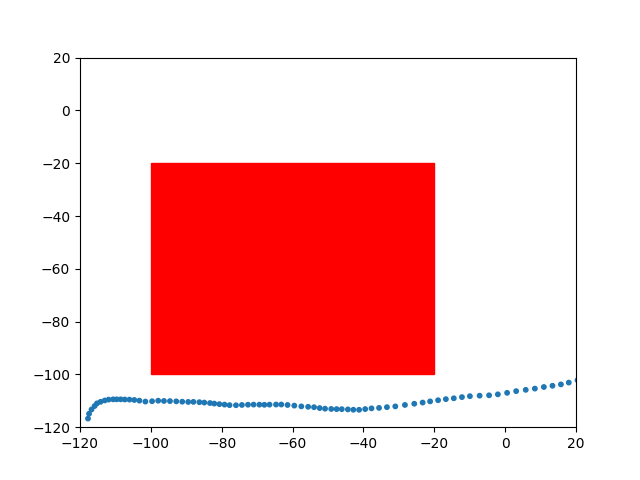}
    \caption{Each row represents two policies trained to reach the goal at $(0, 0)$ from the starting state of $(-120, -120)$. \textbf{Top: $\delta=0$}, where both policies coincidentally choose to take different paths to the goal. This would likely not be the case in multiple separate runs. \textbf{Middle: $\delta=0.01$}, where both agents take different paths to the goal yet again, likely bolstered by the fact that there is now a diversity term between them. \textbf{Bottom: $\delta=0.1$}, which is too high, and surprisingly, both agents take the same route but then diverge after the corner of the obstacle. This experiment shows that an intermediate value of $\delta$ is the most likely to work well.}
    \label{fig:pbot}
\end{figure}

We attempted diversity-promoting pretraining on the Pointbot environment to better understand what the component policies look like. The results are described in detail in the caption of figure \ref{fig:pbot}.

\subsection{Individual Contributions}
\begin{itemize}
    \item Simeon: Worked on experiments on multiple environments and algorithms including baselines. 
    \item Satvik: Worked on code and experiments for modified environments and hierarchical policy method.
    \item Kaushik: Worked on code and experiments for diverse pretraining and the partially supervised method.
\end{itemize}

\end{document}